\pdfoutput=1

\documentclass[11pt]{article}

\usepackage{coling}

\usepackage{times}
\usepackage{latexsym}
\usepackage{threeparttable}
\usepackage{hyperref}
\usepackage{multirow}
\usepackage{booktabs}
\usepackage{graphicx}
\usepackage{textcomp}
\usepackage{array}
\usepackage{bm}
\usepackage{amsmath}
\usepackage{color, colortbl}

\usepackage[T1]{fontenc}

\usepackage[utf8]{inputenc}

\usepackage{microtype}

\usepackage{inconsolata}

\usepackage{graphicx}

\usepackage{times}
\usepackage{latexsym}

\usepackage[T1]{fontenc}

\usepackage[utf8]{inputenc}

\usepackage{microtype}

\usepackage{inconsolata}

\usepackage{times}
\usepackage{latexsym}
\usepackage{balance}
\usepackage[T1]{fontenc}

\usepackage[utf8]{inputenc}

\usepackage{microtype}

\usepackage{inconsolata}
\usepackage{CJKutf8}
\usepackage{booktabs}
\usepackage{graphicx}
\usepackage{array}
\usepackage{bm}
\usepackage{amsmath}
\usepackage{enumitem}
\usepackage{threeparttable}
\usepackage{multirow}
\usepackage{makecell}

\usepackage{latexsym} 

\usepackage[T1]{fontenc}

\usepackage[utf8]{inputenc}
\usepackage{float}

\usepackage{microtype}

\usepackage{xcolor}

\def\BibTeX{{\rm B\kern-.05em{\sc i\kern-.025em b}\kern-.08em
    T\kern-.1667em\lower.7ex\hbox{E}\kern-.125emX}}
\usepackage{cite}
\usepackage{amsmath,amssymb,amsfonts}
\usepackage{algorithmic}
\usepackage[ruled,linesnumbered,noline,noend]{algorithm2e}

\SetCommentSty{mycommfont}

\usepackage[medium,compact]{titlesec}

\SetNlSty{}{}{}

\let\oldnl\nl
\newcommand\nonl{%
  \renewcommand{\nl}{\let\nl\oldnl}}
  
\usepackage{hyperref}
\hypersetup{
    colorlinks=true,
    linkcolor=blue,}
\AtBeginDocument{%
  \providecommand\BibTeX{{%
    \normalfont B\kern-0.5em{\scshape i\kern-0.25em b}\kern-0.8em\TeX}}}

\usepackage{textcomp}
\usepackage{xcolor}
\def\BibTeX{{\rm B\kern-.05em{\sc i\kern-.025em b}\kern-.08em
    T\kern-.1667em\lower.7ex\hbox{E}\kern-.125emX}}
\usepackage{amsmath,amssymb,amsfonts}
\usepackage{algorithmic}
\usepackage[ruled,linesnumbered,noline,noend]{algorithm2e}
\usepackage{threeparttable}
\usepackage{booktabs}
\usepackage{graphicx}
\usepackage{textcomp}
\usepackage{array}
\usepackage{bm}
\usepackage{amsmath}
\usepackage{color, colortbl}
\SetCommentSty{mycommfont}

\usepackage[misc]{ifsym}
\usepackage{amsthm}
\newtheorem{Problem definition}{Problem definition}

\def\BibTeX{{\rm B\kern-.05em{\sc i\kern-.025em b}\kern-.08em
    T\kern-.1667em\lower.7ex\hbox{E}\kern-.125emX}}
\usepackage{hyperref}
\hypersetup{
    colorlinks=true,
    linkcolor=blue,}

\usepackage{color, colortbl}
\usepackage{array}
\usepackage{bm}
\usepackage{amsmath}
\usepackage{threeparttable}
\usepackage{booktabs}
\usepackage{graphicx}
\usepackage{hyperref}
\usepackage{multirow}
\usepackage[ruled,linesnumbered,noline,noend]{algorithm2e}
\usepackage{makecell}
\hypersetup{
    colorlinks=true,
    linkcolor=blue,}
\usepackage{CJKutf8}
\usepackage{booktabs}
\usepackage{graphicx}
\usepackage{textcomp}
\usepackage{array}
\usepackage{bm}
\usepackage{amsmath}

\usepackage{threeparttable}
\usepackage{multirow}
\usepackage{makecell}
\usepackage{times}
\usepackage{latexsym}
\usepackage[T1]{fontenc}

\usepackage[utf8]{inputenc}
\usepackage{float}

\usepackage{microtype}

\usepackage{xcolor}

\def\BibTeX{{\rm B\kern-.05em{\sc i\kern-.025em b}\kern-.08em
    T\kern-.1667em\lower.7ex\hbox{E}\kern-.125emX}}
\usepackage{cite}
\usepackage{amsmath,amssymb,amsfonts}
\usepackage{algorithmic}
\usepackage[ruled,linesnumbered,noline,noend]{algorithm2e}

\SetCommentSty{mycommfont}

\usepackage{fontawesome}

\SetNlSty{}{}{}

\let\oldnl\nl

\usepackage{hyperref}
\hypersetup{
    colorlinks=true,
    linkcolor=blue,}
\AtBeginDocument{%
  \providecommand\BibTeX{{%
    \normalfont B\kern-0.5em{\scshape i\kern-0.25em b}\kern-0.8em\TeX}}}
\usepackage[medium,compact]{titlesec}

\usepackage{times}
\usepackage{latexsym}

\usepackage[T1]{fontenc}

\usepackage[utf8]{inputenc}

\usepackage{microtype}
\usepackage{mathrsfs}

\title{Recent Advancement of Emotion Cognition in Large Language Models}

\author{Yuyan Chen$^{1}$, Yanghua Xiao$^{1}$ $^{\textrm{\Letter}}$ \\
        $^1$ Shanghai Key Laboratory of Data Science, School of Computer Science, Fudan University \\
        \texttt{\{chenyuyan21@m., shawyh@\}fudan.edu.cn}\\
}

\begin{document}
\maketitle
\begin{abstract}
Emotion cognition in large language models (LLMs) is crucial for enhancing performance across various applications, such as social media, human-computer interaction, and mental health assessment. We explore the current landscape of research, which primarily revolves around emotion classification, emotionally rich response generation, and Theory of Mind assessments, while acknowledge the challenges like dependency on annotated data and complexity in emotion processing. 
In this paper, we present a detailed survey of recent progress in LLMs for emotion cognition. We explore key research studies, methodologies, outcomes, and resources, aligning them with Ulric Neisser's cognitive stages. Additionally, we outline potential future directions for research in this evolving field, including unsupervised learning approaches and the development of more complex and interpretable emotion cognition LLMs. We also discuss advanced methods such as contrastive learning used to improve LLMs' emotion cognition capabilities. 
\end{abstract}

\section{Introduction}
In today's emotion computing field, the significance of emotion cognition in large language models (LLMs) is increasingly recognized~\citep{ren2024survey}. It offers profound insights into the complex processes of human emotions and cognition. This area involves not only analyzing the emotional states of individuals or groups but also effectively utilizing these emotions in various applications, such as social media analysis~\citep{chen2024hotvcom,yang2024augmentation,chen2024xmecap,jin2023predicting,jin2024mm}, human-computer interaction~\citep{chen2023xmqas,chen2023can}, and mental health assessment~\citep{chen2024emotionqueen}. Having emotion cognition capabilities enables LLMs to align more closely with human values, thereby enhancing their performance in emotion-related downstream tasks.

\begin{figure}[!t]
  \centering
  \includegraphics[width=0.9\linewidth]{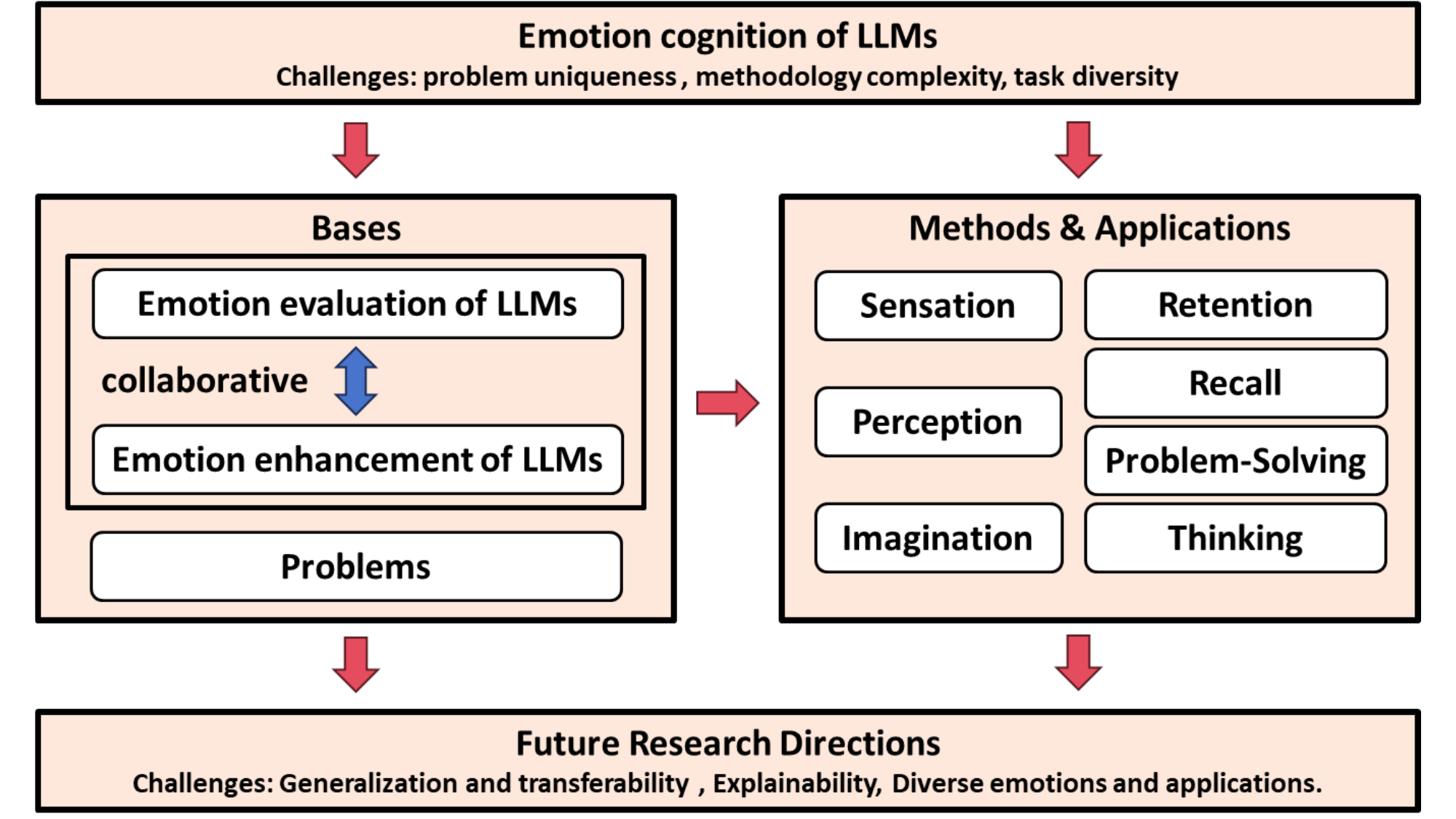}
  \caption{The framework of this survey, including the challenges of LLMs' emotion cognition, Bases and Methods to enhance LLMs' emotion cognition of LLMs, as well as the future direction in this topic.}
  \label{fig:intro}
\end{figure}

Current research in emotion cognition with LLMs focuses on a range of approaches for processing and analyzing emotional data. This includes emotion classification \citep{zhang2023refashioning}, generating emotionally rich responses \citep{xie2023next,chen2024temporalmed}, and Theory of Mind assessments \citep{sap2022neural}. The researchers also focus on enhancing LLMs' emotional capabilities through techniques such as in-context learning \citep{sun2023rational,chen2024drAcademy} and fine-tuning methods \citep{peng2023customising,chen2023mapo}, etc. However, challenges still exist, including over-reliance on annotated data, difficulty in handling complex emotions, and explaining the decision-making processes of LLMs in emotion cognition.
Moreover, 
emotion cognition is highly relevant to human emotional psychology, requiring not only computational methods and techniques but also a deep understanding and application of psychological theories. 

In our review, we emphasize combining a psychological perspective, particularly based on Ulric Neisser’s cognitive psychology theory~\citep{neisser2014cognitive}, to examine the application and research of LLMs in emotion cognition. Neisser, known as the father of cognitive psychology, describing cognition in his book ``Cognitive Psychology'' as a comprehensive process that includes sensation, perception, imagination, memory, recall, problem-solving, and thinking. 
Fig.~\ref{fig:intro} illustrates the framework of this survey.   
Specifically, we first delve into three key challenges faced in emotion cognition of LLMs: the uniqueness of emotional problems, the complexity of emotional methodologies, and the diversity of emotional tasks. We then introduce two typical directions in LLMs for emotion cognition: emotion evaluation and emotion enhancement. Drawing upon the cognitive definition by Ulric Neisser, 
we categorize works in the emotion domain of LLMs into these seven stages. In each stage, we explore more detailed research directions and application scenarios. Additionally, we summarize significant works in emotion cognition, categorized by these seven stages, including their motivation, key methodologies, performance, and available open-source codes and datasets. Lastly, we identify and discuss open questions and potential future directions in the field of emotion cognition of LLMs. The main contributions of our work include:
\begin{itemize}
\item We make an in-depth analysis of the main challenges in emotion cognition from the perspectives of problem definition, methodology, and application areas.
\item We categorize works in the emotion domain according to the seven stages of cognitive psychology theory, aligning specific tasks better with human cognitive processes.
\item We provide an insightful discussion on future research directions in emotion cognition, aiming at inspiring further advancements in the field of emotion computing of LLMs.
\end{itemize}

\section{Challenges and Bases}




Emotion cognition in LLMs faces key challenges: \emph{i) Uniqueness of the Problem,} which indicates emotions in text are abstract and require deep understanding; \emph{ii) Methodological Complexity,} which indicates emotions extends beyond classification or generation, needing tailored strategies and quantitative evaluations; \emph{iii) Diversity of Tasks,} which indicates emotion tasks like question answering and dialogue need varied methods and effectiveness measures. 

We first introduce the bases of emotion cognition, which include evaluations and enhancement for LLMs.
Specifically, existing evaluations for LLMs can be categorized as follows: \emph{i) Discriminative Tasks} include sentiment classification and emotion recognition on various datasets \citep{zhang2023refashioning, carneros2023comparative, rathje2023gpt}. \emph{ii) Generation Tasks} assess LLMs' emotional generation capabilities \citep{lynch2023structured, xie2023next, Klapach2023comparative}. \emph{iii) Theory of Mind} focus on LLMs' abilities to infer and replicate human mental states \citep{trott2023large, sap2022neural, gandhi2023understanding}. \emph{iv) Higher-order Tasks} include role-playing, humor understanding, empathy generation, and decision-making \citep{jiang2023personallm, jentzsch2023chatgpt, lee2022does, jin2022make}.
Enhancement methods for LLMs are categorized as follows: \emph{i) In-context Learning} incorporates emotional labels and contextual examples to improve emotion recognition \citep{sun2023rational, lei2023instructerc, lee2022does}. \emph{ii) Fine-tuning} enhance performance in tasks like mental health prediction and sentiment analysis \citep{xu2023leveraging, kheiri2023sentimentgpt, peng2023customising, binz2023turning}.  \emph{iii) Knowledge Enhancement} is adopted in tasks like completing dialogues, predicting emotional distribution, and integrating with knowledge bases enrich LLMs' emotional content \citep{zheng2023augesc, gagne2023inner, sun2023rational, jeong2023chatbot, qian2023harnessing, shao2023character}.

Next, we elaborate three typical emotion cognition problems with more detailed symbolic representations as follows:
\emph{i) Emotion classification.} Given a dataset \(D = \{ (x_i, y_i) \}_{i=1}^{N}\), where \(x_i\) represents the input text and \(y_i\) is the corresponding labeled emotion, the objective is to develop an LLM, denoted as \(LLM_{EAC}\). This model is tasked with function \(f_{EAC}: X \rightarrow E\), where \(X\) is the space of input texts and \(E\) is the set of possible emotion classes. The goal is for \(LLM_{EAC}\) to predict the emotion class \(e \in E\) for a new input text \(x \in X\), using a limited set of labeled examples \((x_i, y_i)\).
\emph{ii) Emotion generation.} Given a conversational context or narrative \(C = \{c_1, c_2, ..., c_m\}\), where \(c_j\) represents the \(j\)-th element of the context, the task is to develop an LLM, denoted as \(LLM_{EIG}\). The model should implement a function \(f_{EIG}: C \rightarrow R\), where \(R\) is the space of possible responses or generated text. The model \(LLM_{EIG}\) is expected to generate a response \(r \in R\) that is contextually and emotionally aligned with the given conversational context \(C\).
\emph{iii) Emotion interpretability.} Given a set of emotional data \(E = \{e_1, e_2, ..., e_k\}\), where each \(e_k\) represents an annotated emotional instance, the objective is to develop an LLM, denoted as \(LLM_{EI}\). This model should execute a function \(f_{EI}: E \times X \rightarrow I\), where \(I\) is the set of interpretable explanations. The model \(LLM_{EI}\) is designed to provide an explanation \(i \in I\) for the predicted or generated emotion \(e_k\) corresponding to an input text \(x \in X\), elucidating the rationale behind its emotional inference or generation.

\section{Methods and Applications}
In this section, drawing parallels between Neisser's cognitive processes and LLMs' capabilities, we elaborate emotion cognition of LLMs in the same seven stages. we list representative studies with their motivations, key techniques, results, open-source codes/datasets are shown in Table~\ref{tab:paperlist}. The enlarged version is shown in Table~\ref{tab:paperlist1} and Table~\ref{tab:paperlist2}.

\begin{table*}[]
\centering
\resizebox{\textwidth}{!}{%
\begin{tabular}{p{2.5cm}|p{4.5cm}|p{12cm}|p{10cm}|p{21cm}|p{2cm}|p{5cm}}
\toprule
Stage                             & Paper                                                & Motivation                                                                                                                              & Approach                                                                                                        & Performance                                                                                                                        & Venue                                           & Code/Datalink                                                                         \\ \midrule
                                  & \citep{lynch2023structured}          & Generating narratives based on simulated agents.                                                                                        & Structured narrative prompt                                                                                     & ChatGPT's sentiment scores are similar to real tweets in 4 out of 44 categories.                           & Future Internet                                 & https: //data.mendeley.com
                                  /datasets/nyxndvwfsh/2                                      \\
                                  & \citep{ratican2023six}               & Analyzing human emotions within LLMs from multiple angles.                                                  & The Six Emotional Dimension (6DE) model                                                                         & Performance is improved by integrating 6DE model and reinforcement learning.                                                                                                                                   & IJAIRR &                                                                                       \\
                                  & \citep{zhang2023refashioning}        & Exploring the potential of LLMs for emotion recognition.                                                          & Prompt engineering                                                                                              & LLMs use context to improve emotion estimation.                                & arXiv                                           &                                                                                       \\
                                  & \citep{xu2023leveraging}             & Evaluating and improving LLMs' performance in mental health.                                                                                             & Zero-shot prompting, Few-shot prompting, and Instruction fine-tuning                                            & Best-finetuned models outperform GPT-3.5's best prompt by 10.9\% in balanced accuracy, and GPT-4 by 4.8\%.                                                  & arXiv                                           &                                                                                       \\
                                & \citep{binz2023turning}              & Investigating transforming LLMs into cognitive models.                                                                                    & Fine-tuning (on data from psychological experiments)                                                            & CENTaUR achieves -48002.3 negative log-likelihood on the Choices13k dataset and -25968.6 on the Horizon task, beating domain-specific models.                     & arXiv                                           &                                                                                       \\
       \multirow{-7}{*}{Sensation}                              & \citep{sun2023rational}              & Making LLMs better reflect human behavior in conversations.                               & Contextual rationality segregation with attention mechanism                                                     & Lamb excels in generating empathetic responses with an accuracy of 53.44\%.                                                                   & arXiv                                           &                                                                                       \\ 
      & \citep{rathje2023gpt}                & How LLMs represent sentiments and emotions in generated sentences.                                                & Precise sentiment distribution modeling with LLMs                                                               &      GPT-4 outperforms dictionary-based text analysis (r = 0.66-0.75 vs. r = 0.20-0.30 for correlations with manual annotations).                                                                                                                        & arXiv                                           &                                                                                       \\\midrule
                                  & \citep{gagne2023inner}               & Evaluating GPT as a tool for automated psychological text analysis.                                                                       & Multilingual psychological text analysis with GPT                                                              & LLMs predict emotions with less than 1\% deviation for valence and 2-7\% for other emotions.                              & PsyArXiv                                        & https://osf.io/6pnb2                                    \\
                                  & \citep{lei2023instructerc}           & Enhancing LLMs' emotion recognition in the conversations.                                & InstructERC                                                                                                     & InstructERC outperforms UniMSE, SACL-LSTM, and EmotionIC on the IEMOCAP, MELD, and EmoryNLP datasets, respectively, by margins of 0.73\%, 2.70\%, and 1.36\%.                                                            & arXiv                                           &                                                                                       \\
                                 & \citep{peng2023customising}          & Improving emotion recognition with LLMs.                                                                & Adaptation of LLMs for emotion recognition                                                                      & LLMs outperform specialized deep models.                                                                                   & arXiv                                           &                                                                                       \\
      \multirow{-7}{*}{Perception}                               & \citep{kheiri2023sentimentgpt}       & Thoroughly examining various GPT methodologies in sentiment analysis                                                                      & GPT-based sentiment analysis                                                                                    & Achieving over 22\% higher F1-score than the current SOTA methods.                                               & arXiv                                           & https://github.com
                                  \/DSAatUSU
                                  \/SentimentGPT                                              \\
                                  & \citep{trott2023turing}              & Exploring how humans recognize and understand humorous utterances                                                                         & LLM-based humor comprehension                                                                        & In the humor comprehension task, GPT-3's accuracy is lower for jokes (69.4\%) compared with literal statements (expected: 96.2\%, straight: 93.1\%).                      & PsyArXiv                                        &                                                                                       \\
                                  & \citep{ullman2023large}      & Iinvestigating how minor variations affect LLMs' ToM performance.                                      & Minor task alterations                                                                                   &      GPT-3.5's accuracy drops from 99\% to below 50\% with minor modifications                            & arXiv  &   \\
    & \citep{kwon2022representations}      & Exploring methods for representing similarity among emotion concepts.                                      & Emotion representation fusion                                                                                   &      GPT-3 shows a correlation much closer to appraisal feature-based similarity compared with word2vec models.                                                                                                                              & CogSci  & https://doi.org
\/10.7910/DVN
\/6DPPKH                       \\\midrule
                                  & \citep{xie2023next}                  & Comparing the story generation capacity of LLMs                                                                                           & Prompt engineering                                                                                              & GPT-3 generates high-quality stories comparable to those written by humans, outperforming SOTA methods.                                                            & SIGDIAL                                  &                                                                                       \\
                                  & \citep{yongsatianchot2023s}          & Exploring the capabilities of GPT-4 to solve tasks related to emotion prediction                                                               & Emotion prediction and manipulation                                                                             & Prompting GPT-4 to identify key emotional factors enables it to manipulate the emotional intensity of its stories.                                                                                   & ACIIW                                           &                                                                                       \\
                                  & \citep{zheng2023augesc}              & Expanding emotional support dialogue corpora with LLMs.                                            & Language-model augmented dialogue completion                                                                    & AUGESC achieves high scores in informativeness, understanding, helpfulness, consistency, and coherence, based on human evaluation, outperforming strong baselines and crowdsourced dialogues.                             & ACL                                        &                                                                                       \\
                                  & \citep{lee2022does}                  & Exploring whether GPT-3 can generate empathetic dialogues                                                                                 & Prompt-based in-context learning for empathetic dialogue generation                                             & INTENTACC 82.5\%, EMOACC 34.2\% in single-turn setting, and INTENTACC 81.3\%, EMOACC 33.7\% in multi-turn setting.                                                                  & COLING                                          & https://github.com
                                  \/passing2961
                                  \/EmpGPT-3                  \\
                              & \citep{zhao2023chatgpt}              & Evaluating ChatGPT's emotional dialogue capability                                                                                        & Emotional dialogue understanding and generation evaluation method                                               & ChatGPT generally has a performance gap of 3-18\% in ERC, 11.95\% in CEE, and 11-17\% in DAC.                                                              & arXiv                                           &                                                                                       \\
           \multirow{-9}{*}{Imagination}                            & \citep{chen2024talk}       & Enhancing LLMs' understanding and generation of humorous responses in dialogues.                                                                      & Training LMs with a Chinese humor response dataset, chain-of-humor annotations and auxiliary tasks                                                                                    &  Improve LMs' humorous response accuracy by up to 30\% compared to baseline ones                                               & AAAI 2024                                           &                                               \\
                                  & \citep{jentzsch2023chatgpt}          & Evaluating ChatGPT's ability to understand and replicate human humor.                                                              & LLM-based humor understanding and generation                                                                    & ChatGPT's 1008 jokes are too similar to 25 specific jokes, and ChatGPT provides invalid explanations for some jokes.                                                                               & arXiv                                           & https://github.com
                                  \/DLR-SC/JokeGPT-WASSA23                \\
                                  & \citep{toplyn2023witscript}          & Enhancing AI joke generation and develop chatbot with humanlike humor.                        & Witscript 3                                                                                                     & Witscript 3's responses to input sentences have 44\% jokes.                                                           & arXiv                                           &                                                                                       \\
    & \citep{venkatakrishnan2023exploring} & Exploring using transformer models for emotion detection.                                            & Emotion detection task                                                                                              & RoBERTa beats GPT3.5 in emotion detection with fine-tuning.                                        & CAI                                        &                                                                                       \\\midrule
                                  & \citep{tao2023rolecraft}             & Enhancing personalized role-playing with LLMs                                                                            & Personalized character-driven language modeling                                                                 & The RoleCraft-GLM model excels in specific role knowledge and memory with a score of 0.3573 but is slightly less efficient in general instruction response with a score of 0.5385.               & arXiv                                           & https://github.com
                                  \/tml2002/RoleCraft                     \\
                                  & \citep{shao2023character}            & Studying how LLMs can embody individuals.                                        & Character-LLM                                                                                                   & Simulating characters and memorizing experiences effectively.                                                      & arXiv                                           & https://github.com
                                  \/choosewhatulike
                                  \/trainable-agents      \\
                                  & \citep{jiang2023personallm}          & Assessing LLMs' accuracy and consistency in reflecting specific traits.                                                       & Personality-aligned LLM generation                                                                              & Correlate language features with assigned personality types effectively.                 & arXiv                                           &                                                                                       \\
\multirow{-4}{*}{Retention}       & \citep{Klapach2023comparative}       & Exploring emotional abilities of LLMs.                            & Emotional analysis of LLMs                                                                     & GPT4 scores the highest at 889.5/1000, while Bing AI scores the lowest at 427.8/1000 among the 5 LLMs for emotional capabilities.                                          & research archive of rising scholars             &                                                                                       \\\midrule
                                  & \citep{jia2023knowledge}             & Addressing challenges in emotional support conversations.                                                                                  & MODERN                                                                                                          & MODERN records the lowest PPL at 14.99, highest BLEU-1 score at 23.19, BLEU-2 at 10.13, BLEU-3 at 5.53, BLEU-4 at 3.39, ROUGE-L at 20.86, METEOR at 9.26, and the highest CIDEr score at 30.08, outperforming all SOTA baselines on the ESConv dataset.                                                           & arXiv                                           &                                                                                       \\
                                  & \citep{jeong2023chatbot}             & Enhancing realism and consistency of responses from LLMs                          & Multi-sensory contextual response generation                                                                    & Model with all components achieves improved performance, scoring 0.118 in METEOR and 0.209 in Sentence BERT.                                                              & arXiv                                           & https://github.com
                                  \/srafsasm/InfoRichBot                  \\
                                  & \citep{zhong2023memorybank}          & Improving long-term memory in LLMs                                                         & MemoryBank                                                                                                      & Exhibiting strong capability for long-term companionship.                                         & arXiv                                           &                                                                                       \\
     \multirow{-5}{*}{Recall}                              & \citep{qian2023harnessing}           & Exploring the performance of LLMs in generating empathetic responses                                                                      & Semantically similar in-context learning, two-stage interactive generation, and combination with knowledge base & With ChatGPT using 5-shot in-context learning, the Dist-1 and Dist-2 scores reach 2.96 and 18.29, and BLEU-4 score is 2.65.                                                                                                 & arXiv                                           &                                                                                       \\
         & \citep{wake2023bias}                 & Assessing ChatGPT's performance in recognizing emotions from text.                                                                         & Emotion recognition with ChatGPT                                                                                & Emotion datasets vary, indicating potential bias and instability in LLMs.                                         & arXiv                                           &                                                                                       \\\midrule
                                  & \citep{tu2023characterchat}          & Providing emotional support to individuals with diverse personalities & Persona-driven social support conversation                                                              & The results show high effectiveness, with emotional improvement averaging 4.37, problem solving) at 3.30, and active engagement at 4.89 on a 5-point scale.  & arXiv                                           & https://github.com
                                  \/morecry/CharacterChat                 \\
                                  & \citep{qi2023supervised}             & Identifying cognitive distortions and suicidal thoughts.                                  & Psychological risk classification with LLMs                                                   & The fine-tuned GPT-3.5 achieves an F1 score of 78.45\%, an improvement of 11.5\% over the base model.                                                & research square                                 & https://github.com
                                  \/hongzhiQ
                                  \/SupervisedVsLLM
                                  \-EfficacyEval \\
       \multirow{-5}{*}{Problem-Solving}                           & \citep{zhu2024reading}      &  Enhancing LLMs' empathetic understanding to infer users' underlying goals and psychological needs.                                      &  Prompt engineering.                                                                                   &      reaching up to 85\% accuracy in inferring users' goals and fundamental psychological needs                                                                                                                              & arXiv       &                \\
& \citep{sajja2023artificial}          & Creating an smart assistant for personalized learning in higher education.                                 & AI-enabled Intelligent Assistant (AIIA)                                                    & The framework shows enhanced personalized and adaptive learning.                                   & arXiv                                           &                                                                                       \\
& \citep{lai2023psy}                   & Addressing the increasing demand for mental health support.                             & Psy-LLM                                                                                                         & The framework generates coherent and relevant answers effectively.                                                  & arXiv                                           &                                                                                       \\\midrule
                                  & \citep{trott2023large}               & Studying the impact of language exposure on understanding mental states.                                              & Language exposure hypothesis testing                                                                            & GPT-3's predictions are sensitive to characters' knowledge states, and the accuracy reaches 74.5\% and 73.4\% in two evaluations.                                                                                  & Cognitive Science                               & https://osf.io/hu865/                                    \\
                                  & \citep{gandhi2023understanding}      & Investigate the ToM reasoning capabilities of LLMs                                                                     & Causal template-based evaluation framework                                                                      & GPT4 mirrors human ToM, while other LLMs struggle in comparison.                                             & arXiv                                           & https://osf.io/qxj2s                                     \\
                                  & \citep{sap2022neural}                & Exploring social intelligence and ToM in modern NLP systems.                                                      & Instruction-tuned, RLFH, neural ToM                                                                             & GPT-4 performs only 60\% accuracy on the ToM questions                                       & arXiv                                           &                                                                                       \\
        \multirow{-12}{*}{Thinking}  
        & \citep{lee2024large}            &  Investigating LLMs' abilities to generate empathetic responses                                                                                       & Human evaluation                                                                                            & GPT-4 achieves an average empathy score of 4.1 out of 5                                  & arXiv                                            &        \\
                                  & \citep{shapira2023clever}            & Evaluating the extent of N-ToM capabilities in LLMs                                                                                       & N-ToM evaluation                                                                                            & LLMs have ToM abilities but lack robust performance.                                  & arXiv                                           & https://github.com
                                  \/salavi/Clever\_Hans\_or
                                  \_N-ToM        \\
                                  & \citep{zhou2023far}                  & Evaluating LLMs' ability to link mental states with social actions.                                        & Thinking for Doing (T4D), Zero-shot prompting framework, Foresee and Reflect (FaR)                                                                                        & FaR boosts GPT-4's performance from 50\% to 71\% on T4D                                         & arXiv                                           &                                                                                       \\
                                  & \citep{sorin2023large}      &  Reviewing LLMs' empathy in healthcare.                                      &  A literature search on MEDLINE                                                                                   &      ChatGPT-3.5 outperforms humans in empathy-related tasks, achieving up to 85\%                                                                                                                              & MedRxiv  &                 \\
                                    & \citep{schaaff2023exploring}      &  Exploring ChatGPT's ability to understand and express empathy                                      &  A series of tests and questionnaires                                                                                   &      ChatGPT accurately identified emotions in 91.7\% of cases and produced parallel emotional responses in 70.7\% of conversations                                                                                                                              & arXiv  &                 \\
                                    & \citep{saito2023chatgpt}      &  Enhancing LLMs' empathetic dialogue speech synthesis                                       &  Use ChatGPT to extract context words and train an empathetic speech synthesis model.                                                                                   &      ChatGPT-EDSS achieves 3.52 for naturalness and 3.24 for speaking style similarity                                                                                                                              & INTERSPEECH 2023  & 
                                    \\
                                    
& \citet{holterman2023does} & understand whether ChatGPT-3 and ChatGPT-4 can exhibit ToM. & Using six well-known problems that focus on human reasoning biases and decision-making & ChatGPT-4 provides 30\% more correct answers than ChatGPT-3 even if it's based on incorrect assumptions or reasoning. &arXiv&\\
       & \citep{jin2022make}                  & Predicting human moral judgments for AI collaboration and safety.                                           & Moral Chain-of-Thought (MoralCoT) prompting strategy                                                            & MoralCoT outperforms existing LLMs by 6.2\% F1.                                                                   & NeurIPS                                    & https://huggingface.co
\/datasets/feradauto
\/MoralExceptQA  \\ 
& \citet{del2022empathy} & Enhancing GPT-3' ability to detect empathy in text. & Transfer learning and fine-tuning & Achieve an F1 score of 0.73 on the empathy detection task. &WASSA 2022&\\
\bottomrule 
\end{tabular}%
}
\caption{A list of representative papers among each level with motivation, approach, performance, venue, and code/datalink.}
\label{tab:paperlist}
\end{table*}

\subsection{Sensation}
Sensation is the notion that LLMs exhibit capabilities akin to human in processing input textual data. Work in this area mainly focuses on the input form.
The common input forms contain three categories, including prompt engineering, embeddings representation and knowledge enhancement. 

Prompt engineering means adding some instructions to guide LLMs on downstream tasks. 
For example, \citet{lynch2023structured}  
introduced a structured narrative prompt designed for querying LLMs. The study uses OpenAI's ChatGPT to generate narratives, then compares the emotional levels in these narratives with real tweets using statistical tests like chi-squared and Fisher's exact tests; 
%
\citet{ratican2023six} proposed the 6DE model to analyze human emotions in LLM contexts. This model considers multiple dimensions of emotions, such as arousal, valence, dominance, agency, fidelity, and novelty; 
\citet{zhang2023refashioning}
explored four prompt strategies, including zero-shot and few-shot prompts with and without context, demonstrating the good performance of these prompts for LLMs in emotion analysis and recognition tasks comparing with those without any prompting strategy. The study highlights the importance of contextual information in enhancing emotion estimation by LLMs.

Different from using prompts for input, an embeddings representation converts the input text into a vector representation in a high-dimensional space, capturing the semantic information of the vocabulary, which is used for optimizing internal process of the LLM.
For example, \citet{xu2023leveraging} explored instruction tuning to enhance LLM performance in mental health prediction. The fine-tuned models, Mental-Alpaca and Mental-FLAN-T5, notably surpass GPT-3.5 and GPT-4's performance, despite being significantly smaller in size;
\citet{binz2023turning} investigated psychological experiment data which is used to fine-tune LLMs. This research demonstrates LLMs' capabilities in accurately mimicking human behavior, and indicating LLMs' potential in emotion cognition with embeddings representation in the fine-tuning process.

Moreover, knowledge enhancement means adding context or knowledge into the input to enhance LLMs' performance in processing downstream tasks. 
For example, \citet{sun2023rational} focused on enhancing empathetic response generation by incorporating external knowledge. This study introduces a novel approach called CoNECT, which utilizes emotional indicators to assess context relevance and promote empathy reasoning; 
\citet{gagne2023inner} explored the sentiment distribution of text generated by LLMs. 
This approach enables the generation of emotionally rich sentences through the utilization of specific quantiles, demonstrating LLMs' effectiveness in emotion-related generation, and offers insights into LLMs' internal mechanism.

In brief, current research in LLMs has achieved notable progress in processing emotional text inputs, \emph{mainly through prompt engineering, embeddings representation, and knowledge enhancement.} These methods have enhanced LLMs' capabilities to \emph{understand and generate emotion-rich content.} However, there remains room for improvement in \emph{diversifying the modes of receiving and processing text inputs}, including integrating a more nuanced internal emotional cognition within LLMs to better interpret and respond to inputs.

\subsection{Perception}
Perception involves interpreting and understanding sensory information, processing raw data collected from the senses to form a meaningful understanding of the external world. 
LLMs' perception in emotion cognition mainly contains emotion recognition and its interpretability.

Emotion recognition involves recognizing emotions in the context or conversations.
For example, \citet{rathje2023gpt} explored the performance of GPT-3.5 and GPT-4 in detecting various language psychological constructs (emotions, discrete emotions, and aggressiveness), suggesting that LLMs are more accurate than dictionary-based methods and fine-tuned machine learning models;
\citet{zhang2023refashioning} demonstrated that LLMs can achieve comparable or superior performance in emotion recognition tasks, especially in identifying minority emotional categories;
\citet{lei2023instructerc} introduced the InstructERC framework, an effective generative framework employing a combination of a retrieval module and an emotion alignment task for emotion recognition; 
\citet{venkatakrishnan2023exploring} emphasized the importance of emotion detection in cross-cultural contexts, examining LLMs' responses to significant events such as the murder of Zhina (Mahsa) Amini in Iran and the earthquake in Turkey and Syria; 
\citet{rodriguez2023review} assessed sentiment analysis methods in social networks and their applications in fields like stock market valuation, politics, and online bullying education. The study finds poor performance using LLMs like GPT-3, and GPT-J, requiring domain-specific adjustments; 
\citet{peng2023customising} adopted deep prompt tuning and low-rank adaptation to investigate how well LLMs perform in linguistic emotion recognition. The impressive performance of the adapted LLMs across six widely used datasets highlights their strong transferability and feasibility in emotion recognition, surpassing other specialized deep models; 
\citet{kheiri2023sentimentgpt} discussed the potential of using LLMs for sentiment analysis, showing that LLMs excel in handling nuances in language for sentiment analysis;
\citet{ullman2023large} emphasized GPT-3.5's skill in predicting human emotions, highlighting its capabilities in understanding and interpreting emotional content in text. 
\citet{carneros2023comparative} conducted a comparative analysis between GPT-3.5 and IBM Watson using a dataset of 30,000 tweets related to the Covid-19 pandemic. This study reveals the multifaceted capabilities of LLMs in sentiment analysis and emotion classification. 
However, they also struggle with fitting textual expressions into defined emotion categories.
Moreover, humor is a more challenging research field in emotion cognition.
\citet{trott2023turing} investigate the ability of GPT-3 to understand verbal humor. Experiments demonstrate that GPT-3 performs above chance in detecting, appreciating, and comprehending jokes, although it does not match human performance. It suggests that while LLMs are adept at grasping humor, language alone is not sufficient for fully getting the joke. Images are also useful.

Interpretability of emotion recognition is to analyze the internal state of LLMs through weight distribution of words, gradient, disturbance. \citet{kwon2022representations} investigated methods of representing emotional concepts by comparing performance between appraisal feature-based and word embedding-based similarity calculation methods. It finds that GPT-3 outperforms in word embedding-based similarity calculations but also relies excessively on the valuation of emotional concepts.

In general, recent advancements in LLMs have focused on improving their perception of emotions in text, mirroring human-like understanding of emotional nuances. While LLMs like \emph{GPT-3.5 and GPT-4 show proficiency in detecting and interpreting emotions across diverse contexts, they still face challenges in fully grasping the context and subtleties of emotions}, highlighting the need for further enhancement in their perceptual capabilities for accurate emotion categorization, depth of understanding, domain adaptability, and value alignment.

\subsection{Imagination}
Imagination is the generation of emotionally relevant content, such as emotional stories, poetry, and emotive dialogues, with the aim of creating content that aligns with human values.

For LLMs' generating emotional narratives, \citet{xie2023next} focused on variations in style, register, and story length in crafting stories, revealing a significant superiority of LLMs in generating story content. However, a critical observation is that LLMs tend to replicate real-world stories when dealing with world knowledge.
\citet{yongsatianchot2023s} investigated the GPT-4's proficiency in tasks associated with emotion prediction, showcasing its capability to not only discern and conceptualize emotion theories but also to create emotion-related stories. By prompting GPT-4 to recognize and manipulate key elements of emotional experiences, it demonstrates a nuanced control over the emotional intensity in its narratives. 

For generating emotive dialogues, \citet{zheng2023augesc} adopted LLMs for dialogue augmentation in emotional support conversations. This approach treats dialogue augmentation as a dialogue completion task, where a fine-tuned language model completes dialogues from various topics, followed by heuristic-based postprocessing. 
\citet{lee2022does} delved into the ability of GPT-3 to generate empathetic dialogues through prompt-based in-context learning. The study introduces innovative context example selection methods, SITSM and EMOSITSM, which leverages emotional and situational information, revealing that GPT-3 achieves competitive performance against Blender 90M in empathy.
\citet{zhao2023chatgpt} examined the emotional dialogue capabilities of ChatGPT, assessing ChatGPT's performance in understanding and generating emotional dialogues through a series of downstream tasks. %
\citet{guo2023can} indicated the emotional effect of ChatGPT in vertical fields such as painting creation. It can provide clearer, more detailed painting instructions, and understand abstract artistic expression and emotion in painting.

We also focus on a more specific aspect: the generation of humor, a complex and inherently human characteristic. 
For example, \citet{jentzsch2023chatgpt} critically investigated OpenAI's ChatGPT in terms of its humor generation capabilities, assessesing ChatGPT's ability to generate, explain, and detect jokes. ChatGPT is expected to repeat the same process of jokes rather than creating new ones, though it can accurately explain valid jokes.
\citet{toplyn2023witscript} presented an innovative approach to humor generation for LLMs. Witscript 3 employs three joke production mechanisms to generate and select the best comedic responses. It represents a collaboration between LLMs and human expertise, incorporating humor algorithms crafted by professional comedy writers. Notably, Witscript 3's responses are perceived as jokes by human evaluators 44\% of the time.
%
\citet{chen2024talk} constructed a Chinese Explainable Humor Response Dataset with chain-of-humor and humor mind map annotations as well as humor-related auxiliary tasks to evaluate and improve PLM and LLMs' humorous response ability of PLMs.

In summary, LLMs have significant \emph{achievements in the imagination of emotionally relevant content, such as stories, dialogues, and humor}. They exhibit capabilities in nuanced control over emotional intensity and empathy in conversations. However, limitations persist in the \emph{originality of content, particularly in humor generation where LLMs tend to replicate existing jokes rather than creating new ones}.

\subsection{Retention}
Retention is the process of encoding and storing knowledge, and creating ``memories''.
Retention in LLMs relates to how they ``remember'' emotional information and knowledge through their training process, which is crucial for subsequent emotional data processing and generation, such as role-playing and character simulation. 

For example, \citet{tao2023rolecraft} introduced an innovative framework aiming at improving personalized role-playing with LLMs. They adopt a detailed emotion classification strategy and annotate emotions within the dialogue dataset, enabling GPT-4 to create character profiles based on emotions in its ``memory''.
\citet{shao2023character} shifted the focus to training agents based on specific personal profiles, experiences, and emotional states, rather than using limited prompts to guide the ChatGPT's API. Experimental results indicate that editing and restoring personal profiles contribute to building simulacra for LLMs that are more accurate and emotionally aware, resembling their characters in a more human-like manner.
\citet{jiang2023personallm} delved into the extent to which the behaviors of personalized LLMs reflect specific personality traits. This study uses the Big Five personality model to create distinct LLM personas and evaluates their behavior through various tasks, including a personality test and story writing. The results reveal that LLM personas can consistently exhibit behaviors aligning with their assigned personality profiles.
\citet{wang2023rolellm} introduced the RoleLLM framework which is used to enhance LLMs' role-playing capacities. This framework includes role outline construction, context-based instruction generation, and role-specific knowledge capturing and retention, showcasing competitive results of LLMs through mimicking linguistic styles and utilizing role-specific knowledge based on their memory.

We also investigated LLMs' capabilities in constructing emotional memory patterns and restoring them. For example, \citet{Klapach2023comparative} delved into the comparative analysis of five widely recognized LLMs, including BingAI~\footnote{https://wwbing.wangdu.site/web}, ChatGPT~\footnote{https://chat.openai.com/}, GoogleBard~\footnote{https://bard.google.com/}, and HuggingChat~\footnote{https://huggingface.co/chat/}, focusing on their ability to process, mimic, and express emotions. These LLMs are tasked with creating new stories that mirror the tone, style, and emotional impact of the original narratives in order to assess their capabilities in storing emotional aspects of the stories and replicating them effectively.
\citet{russo2023countering} introduced a novel method that LLMs manage to restore human annotation with an author-reviewer pipeline combat misinformation on social media platforms by generating emotional responses.

Generally, recent research in LLMs has focused on \emph{enhancing their retention of emotional information, crucial for tasks like role-playing and character simulation}. However, challenges persist in effectively \emph{encoding and storing complex emotional knowledge, being unable to ``retain'' memory due to context size limitations.}. The field explores ways to improve LLMs' memory patterns and their capability to retain emotional information accurately and systematically.

\subsection{Recall}
Recall is the retrieval of emotional memories, extracting ``memories''.
Recall in emotion cognition for LLMs pertains to their capabilities to retrieve emotionally relevant information from their internal or external knowledge in the context of responding emotion-related statement, maintaining consistency of emotional dialogues, etc.
The following studies showcase how LLMs effectively utilize emotional memories and histories to enhance decision-making processes.

For example, \citet{jia2023knowledge} introduced knowledge-enhanced memory model for emotional support conversation. This model is adept at perceiving and adapting to the dynamic emotional shifts within different periods of a conversation through extracting rich knowledge from dialogues and commonsense from ConceptNet~\citep{speer2017conceptnet}. 
\citet{jeong2023chatbot} proposed a novel approach that enriches LLMs' responses by incorporating a diverse set of parameters, including the five senses, attributes, emotional states, relationships with the interlocutor and memories. They underscore the importance of memory in maintaining the continuity and emotional authenticity of conversations.
%
\citet{zhong2023memorybank} introduced a dynamic memory mechanism that makes LLMs utilize past emotional interactions in current decision-making. The mechanism, inspired by the Ebbinghaus Forgetting Curve~\citep{ebbinghaus1885gedachtnis}, allows LLMs to selectively recall emotional interactions, therefore acting more as a real human friend.
\citet{qian2023harnessing} underscored the importance of LLMs' capabilities to generate empathetic responses based on historical emotional contexts. They introduce in-context learning and two-stage interactive generation that enable LLMs to process and reflect upon past emotional interactions, thereby making more empathetically informed decisions. 
In addition, \citet{wake2023bias} investigated the application of emotional history in decision-making. The authors estimate the emotional label of current utterances solely based on the past conversation's history, showcasing the great effect of dataset and emotion label selection on ChatGPT's emotion recognition performance.

In summary, recent studies on recall of LLMs focus on retrieving and utilizing emotional memories for decision-making in emotion-related interactions. These studies \emph{introduce LLMs are adapt to dynamic emotional shifts}, incorporate diverse parameters, and utilize historical emotional contexts for empathetic decision-making. However, challenges remain in \emph{perfecting the recall of complex emotional histories, efficiently retrieving the most relevant knowledge and update it consistently}.

\subsection{Problem-Solving}
Problem-Solving in emotion cognition is to solve emotion-related downstream tasks in various scenarios~\citep{li2023s2phere,chen2024dolarge,li2022meta,li2024vqa,li2023mhrr,ni2024earnings,li2023ltrgcn,li2024gs2p}.

In mental health, for example, \citet{tu2023characterchat} presented the S2Conv framework, tailored for providing personalized support for mental health issues. It integrates personality and memory-based dialogue models with an interpersonal matching plugin, highlighting LLMs' potential in providing social support~\citep{ni2024timeseries,202407.2102,jin2023better,li2024deception}.
\citet{qi2023supervised} evaluated LLMs' performance in the mental health domain. 
\citet{zhu2024reading} investigated the use of LLMs for performing mental inference tasks, specifically inferring users' underlying goals and fundamental psychological needs. 
\citet{lai2023psy} used LLMs in psychological counseling settings, providing immediate responses and mindfulness activities.
\citet{xu2023leveraging} assessed LLMs' performance in mental health prediction tasks, highlighting the need for bias mitigation.
In education, \citet{sajja2023artificial} propose an innovative framework for personalized and adaptive learning.

These diverse applications of LLMs in emotion-related problem-solving highlight their broad potential. However, they also reveal limitations like \emph{gender biases and the need for enhanced interpretability}. Future research directions may \emph{include fine-tuning LLMs to better meet the needs of educational and mental health domains.}

\subsection{Thinking}
Thinking refers to the reflection and review after problem-solving. In the context of emotion cognition in LLMs, it pertains to how emotional Theory of Mind is utilized to solve downstream tasks~\citep{zhou2024reconstruction,202407.0981}.
Theory of Mind (ToM)~\citep{carlson2013theory} is the cognitive capability to understand one's own and others' mental states, including emotions, intentions, expectations, thoughts, beliefs. One can use this theory to predict and interpret various behaviors. 

LLMs have shown promising performance in emotion-driven tasks with powerful thinking capability.
For example, \citet{trott2023large} demonstrated LLMs perform well in inferring others' beliefs. 
\citet{gandhi2023understanding} introduced a causality-based template methodology for evaluating LLMs' ToM. They find that GPT-4 has human-like reasoning patterns with powerful ToM.
\citet{sap2022neural} evaluated GPT-3's performance in social reasoning and mental state comprehension, identifying the boundaries of LLMs' ToM. 
\citet{shapira2023clever} conducted extensive experimentation on six tasks with varied detection methods in assessing LLMs' ToM, which are non-robust and reliant on superficial heuristic methods rather than solid reasoning. 
\citet{holterman2023does} examined the ability of ChatGPT-3 and ChatGPT-4 to exhibit ToM by presenting them with six problems addressing human reasoning biases, finding that ChatGPT-4 provides correct answers more often than chance, albeit sometimes based on incorrect assumptions or flawed reasoning.

In the realm of thinking, the focus extends beyond ToM to behaviors.
For example, 
\citet{zhou2023far} introduced a novel ``Thinking for Doing'' evaluation paradigm, assessesing whether LLMs can discern appropriate actions based on others' mental states, beyond merely responding to questions about these states. The study proposes a zero-shot prompting framework, ``Foresee and Reflect'', to enhance LLMs in predicting future events and reasoning through action choices.
\citet{jin2022make} aimed to evaluate LLMs in comprehending and predicting human moral judgment and decision-making behaviors. They propose a novel moral reasoning chain prompt strategy named MoralCoT based on legal expertise and moral reasoning theories, suggesting that MoralCoT surpasses existing legal models in moral reasoning, complex moral judgments and decisions.
\citet{sorin2023large} reviewed the capacity of LLMs to demonstrate empathy, exploring how LLMs process and express complex emotional viewpoints and reasoning. \citet{del2022empathy} acknowledged GPT-3's role in empathy and distress predictions, highlighting its reasoning process in complex emotional forecasting. \citet{schaaff2023exploring} assessed ChatGPT's empathy levels compared to human standards, comparing the LLMs' understanding and expression of empathy. \citet{saito2023chatgpt} demonstrated the comparable performance of the proposed ChatGPT-EDSS in capturing empathy in dialogues, evaluating the LLMs' emotional understanding and expression capabilities. \citet{lee2024large} found that LLMs' responses are more empathetic than humans', comparing the emotional reasoning and expression between models and humans.

Collectively, these studies have focused on LLMs thinking capability. While LLMs like GPT-4 show promise in \emph{understanding and inferring mental states}, challenges exist in their \emph{depth of reasoning and reflection, better utilizing Theory of Mind, and emotion-driven behavior prediction}. The field aims to enhance LLMs' capabilities in retrospectively analyzing emotional tasks for more nuanced and accurate problem-solving.

\section{Future Research Directions}
Emotion cognition is an emerging and rapidly developing research topic. Despite significant progress, many challenges still exist for future research. 
In this section, we identify and briefly discuss some potential directions.


\begin{itemize}
    \item \textbf{Generalization and transferability.} Current methods in emotion cognition for LLMs primarily rely on labeled data and prompt-based techniques. Notable limitations are the distributional shift between the data used for training and testing, as well as the dependence on the prompter's expertise, which affects the generalization and transferability of these networks~\citep{chen2023hadamard,xiong2024largelanguagemodelslearn,xiao2023large}.
    ~\citet{rathje2023gpt} emphasized the challenges posed by the lack of manually annotated datasets in many under-researched languages, limiting the analysis of GPT's accuracy. 
    \item \textbf{Explainability.}  
    Developing LLMs that provide clear explanations is essential for improving the reliability and trustworthiness of LLMs~\citep{chen2022grow,chen2023hallucination,weng2024leveraging,Weng202404,Weng202406,Weng2024}. 
    For instance, ~\citet{kwon2022representations} delved into how different LLMs perceive and express nuanced human emotions. Additionally, as ~\citet{jentzsch2023chatgpt} pointed out there is a need for deeper understanding beyond just analyzing system outputs. Integrating theories like emotional cognition, as discussed in ~\citep{yongsatianchot2023investigating}, can be a step forward in this direction. Furthermore, ~\citet{yongsatianchot2023s} suggested that while LLMs contribute significantly, they struggle in modeling emotions in physiological, neural, and cognitive aspects.
    \item \textbf{Diverse emotions and applications.} There is a vast array of complex and mixed emotions as well as different modals that have yet to be thoroughly explored~\citep{tao2024nevlp}. ~\citet{trott2023turing} highlighted the growing interest in specific emotions like humor, but other emotions, such as anxiety, remain under-researched. In addition, ~\citet{xu2023leveraging} provided guidelines for enhancing LLMs in mental health prediction tasks. \citet{sajja2023artificial} explored the potential of LLMs in creating more personalized and adaptive learning environments. Lastly, ~\citet{lai2023psy} outlined the limitations and areas for improvement in mental health support models. 

    
\end{itemize}

\section{Conclusion}
Our comprehensive review in this paper highlights the growing importance of emotion cognition in LLMs within the field of emotion computing. We first explore the key challenges in emotion cognition.
Next, we make a categorization of current studies based on Ulric Neisser's cognitive psychology theory, offering a structured summary of emotion processing in LLMs across various stages. This includes analysis and discussion on emotion classification, generation, interpretability, and the integration of these aspects into LLMs to enhance their empathetic and cognitive capabilities.
After that, we summarize motivations, methodologies, results, and available tools of current studies, discussing future directions in emotion cognition of LLMs.
Overall, we contribute to the summary of current trends and future work in emotion cognition of LLMs, expecting to make LLMs be more aligned with human emotion and cognition, therefore better solving emotion-related downstream tasks.
Exploring the emotion cognition capabilities of LLMs can inform further research by improving emotion recognition and response generation, leading to more accurate sentiment analysis and empathetic interactions. This can enhance personalized user experiences in education, entertainment, and customer service in the future.

\section*{Limitations}
While this survey provides a comprehensive overview of recent progress in emotion cognition for LLMs, limitations exist. The primary one is the categorization of research based on Ulric Neisser's cognitive stages may oversimplify the complex nature of emotion cognition in LLMs. Additionally, the focus on key methodologies and outcomes might not capture all nuances of the field. 
In the future, we could explore more nuanced categorizations beyond Ulric Neisser's cognitive stages to provide a more comprehensive and up-to-date understanding of emotion cognition in LLMs.
\bibliography{main}




\begin{table*}[]
\centering
\resizebox{\textwidth}{!}{%
\begin{tabular}{p{2.5cm}|p{3cm}|p{4cm}|p{4cm}|p{7cm}|p{3cm}|p{4cm}}
\toprule
Stage                             & Paper                                                & Motivation                                                                                                                              & Approach                                                                                                        & Performance                                                                                                                        & Venue                                           & Code/Datalink                                                                         \\ \midrule
                                  & \citep{lynch2023structured}          & Generating narratives based on simulated agents.                                                                                        & Structured narrative prompt                                                                                     & ChatGPT's sentiment scores are similar to real tweets in 4 out of 44 categories.                           & Future Internet                                 & https: //data.mendeley.com
                                  /datasets/nyxndvwfsh/2                                      \\
                                  & \citep{ratican2023six}               & Analyzing human emotions within LLMs from multiple angles.                                                  & The Six Emotional Dimension (6DE) model                                                                         & Performance is improved by integrating 6DE model and reinforcement learning.                                                                                                                                   & IJAIRR &                                                                                       \\
                                  & \citep{zhang2023refashioning}        & Exploring the potential of LLMs for emotion recognition.                                                          & Prompt engineering                                                                                              & LLMs use context to improve emotion estimation.                                & arXiv                                           &                                                                                       \\
                                  & \citep{xu2023leveraging}             & Evaluating and improving LLMs' performance in mental health.                                                                                             & Zero-shot prompting, Few-shot prompting, and Instruction fine-tuning                                            & Best-finetuned models outperform GPT-3.5's best prompt by 10.9\% in balanced accuracy, and GPT-4 by 4.8\%.                                                  & arXiv                                           &                                                                                       \\
                                & \citep{binz2023turning}              & Investigating transforming LLMs into cognitive models.                                                                                    & Fine-tuning (on data from psychological experiments)                                                            & CENTaUR achieves -48002.3 negative log-likelihood on the Choices13k dataset and -25968.6 on the Horizon task, beating domain-specific models.                     & arXiv                                           &                                                                                       \\
       \multirow{-7}{*}{Sensation}                              & \citep{sun2023rational}              & Making LLMs better reflect human behavior in conversations.                               & Contextual rationality segregation with attention mechanism                                                     & Lamb excels in generating empathetic responses with an accuracy of 53.44\%.                                                                   & arXiv                                           &                                                                                       \\ 
      & \citep{rathje2023gpt}                & How LLMs represent sentiments and emotions in generated sentences.                                                & Precise sentiment distribution modeling with LLMs                                                               &      GPT-4 outperforms dictionary-based text analysis (r = 0.66-0.75 vs. r = 0.20-0.30 for correlations with manual annotations).                                                                                                                        & arXiv                                           &                                                                                       \\\midrule
                                  & \citep{gagne2023inner}               & Evaluating GPT as a tool for automated psychological text analysis.                                                                       & Multilingual psychological text analysis with GPT                                                              & LLMs predict emotions with less than 1\% deviation for valence and 2-7\% for other emotions.                              & PsyArXiv                                        & https://osf.io/6pnb2                                    \\
                                  & \citep{lei2023instructerc}           & Enhancing LLMs' emotion recognition in the conversations.                                & InstructERC                                                                                                     & InstructERC outperforms UniMSE, SACL-LSTM, and EmotionIC on the IEMOCAP, MELD, and EmoryNLP datasets, respectively, by margins of 0.73\%, 2.70\%, and 1.36\%.                                                            & arXiv                                           &                                                                                       \\
                                 & \citep{peng2023customising}          & Improving emotion recognition with LLMs.                                                                & Adaptation of LLMs for emotion recognition                                                                      & LLMs outperform specialized deep models.                                                                                   & arXiv                                           &                                                                                       \\
      \multirow{-7}{*}{Perception}                               & \citep{kheiri2023sentimentgpt}       & Thoroughly examining various GPT methodologies in sentiment analysis                                                                      & GPT-based sentiment analysis                                                                                    & Achieving over 22\% higher F1-score than the current SOTA methods.                                               & arXiv                                           & https://github.com
                                  \/DSAatUSU
                                  \/SentimentGPT                                              \\
                                  & \citep{trott2023turing}              & Exploring how humans recognize and understand humorous utterances                                                                         & LLM-based humor comprehension                                                                        & In the humor comprehension task, GPT-3's accuracy is lower for jokes (69.4\%) compared with literal statements (expected: 96.2\%, straight: 93.1\%).                      & PsyArXiv                                        &                                                                                       \\
                                  & \citep{ullman2023large}      & Iinvestigating how minor variations affect LLMs' ToM performance.                                      & Minor task alterations                                                                                   &      GPT-3.5's accuracy drops from 99\% to below 50\% with minor modifications                            & arXiv  &   \\
    & \citep{kwon2022representations}      & Exploring methods for representing similarity among emotion concepts.                                      & Emotion representation fusion                                                                                   &      GPT-3 shows a correlation much closer to appraisal feature-based similarity compared with word2vec models.                                                                                                                              & CogSci  & https://doi.org
\/10.7910/DVN
\/6DPPKH                       \\\midrule
                                  & \citep{xie2023next}                  & Comparing the story generation capacity of LLMs                                                                                           & Prompt engineering                                                                                              & GPT-3 generates high-quality stories comparable to those written by humans, outperforming SOTA methods.                                                            & SIGDIAL                                  &                                                                                       \\
                                  & \citep{yongsatianchot2023s}          & Exploring the capabilities of GPT-4 to solve tasks related to emotion prediction                                                               & Emotion prediction and manipulation                                                                             & Prompting GPT-4 to identify key emotional factors enables it to manipulate the emotional intensity of its stories.                                                                                   & ACIIW                                           &                                                                                       \\
                                  & \citep{zheng2023augesc}              & Expanding emotional support dialogue corpora with LLMs.                                            & Language-model augmented dialogue completion                                                                    & AUGESC achieves high scores in informativeness, understanding, helpfulness, consistency, and coherence, based on human evaluation, outperforming strong baselines and crowdsourced dialogues.                             & ACL                                        &                                                                                       \\
                                  & \citep{lee2022does}                  & Exploring whether GPT-3 can generate empathetic dialogues                                                                                 & Prompt-based in-context learning for empathetic dialogue generation                                             & INTENTACC 82.5\%, EMOACC 34.2\% in single-turn setting, and INTENTACC 81.3\%, EMOACC 33.7\% in multi-turn setting.                                                                  & COLING                                          & https://github.com
                                  \/passing2961
                                  \/EmpGPT-3                  \\
                              & \citep{zhao2023chatgpt}              & Evaluating ChatGPT's emotional dialogue capability                                                                                        & Emotional dialogue understanding and generation evaluation method                                               & ChatGPT generally has a performance gap of 3-18\% in ERC, 11.95\% in CEE, and 11-17\% in DAC.                                                              & arXiv                                           &                                                                                       \\
           \multirow{-9}{*}{Imagination}                            & \citep{chen2024talk}       & Enhancing LLMs' understanding and generation of humorous responses in dialogues.                                                                      & Training LMs with a Chinese humor response dataset, chain-of-humor annotations and auxiliary tasks                                                                                    &  Improve LMs' humorous response accuracy by up to 30\% compared to baseline ones                                               & AAAI 2024                                           &                                               \\
                                  & \citep{jentzsch2023chatgpt}          & Evaluating ChatGPT's ability to understand and replicate human humor.                                                              & LLM-based humor understanding and generation                                                                    & ChatGPT's 1008 jokes are too similar to 25 specific jokes, and ChatGPT provides invalid explanations for some jokes.                                                                               & arXiv                                           & https://github.com
                                  \/DLR-SC/JokeGPT-WASSA23                \\
                                  & \citep{toplyn2023witscript}          & Enhancing AI joke generation and develop chatbot with humanlike humor.                        & Witscript 3                                                                                                     & Witscript 3's responses to input sentences have 44\% jokes.                                                           & arXiv                                           &                                                                                       \\
    & \citep{venkatakrishnan2023exploring} & Exploring using transformer models for emotion detection.                                            & Emotion detection task                                                                                              & RoBERTa beats GPT3.5 in emotion detection with fine-tuning.                                        & CAI                                        &                                                                                       \\
\bottomrule 
\end{tabular}%
}
\caption{A list of representative papers among each level with motivation, approach, performance, venue, and code/datalink.}
\label{tab:paperlist1}
\end{table*}

\begin{table*}[]
\centering
\resizebox{\textwidth}{!}{%
\begin{tabular}{p{2.5cm}|p{3cm}|p{4cm}|p{4cm}|p{7cm}|p{3cm}|p{4cm}}
\toprule
Stage                             & Paper                                                & Motivation                                                                                                                              & Approach                                                                                                        & Performance                                                                                                                        & Venue                                           & Code/Datalink                                                                         \\ \midrule
                                  & \citep{tao2023rolecraft}             & Enhancing personalized role-playing with LLMs                                                                            & Personalized character-driven language modeling                                                                 & The RoleCraft-GLM model excels in specific role knowledge and memory with a score of 0.3573 but is slightly less efficient in general instruction response with a score of 0.5385.               & arXiv                                           & https://github.com
                                  \/tml2002/RoleCraft                     \\
                                  & \citep{shao2023character}            & Studying how LLMs can embody individuals.                                        & Character-LLM                                                                                                   & Simulating characters and memorizing experiences effectively.                                                      & arXiv                                           & https://github.com
                                  \/choosewhatulike
                                  \/trainable-agents      \\
                                  & \citep{jiang2023personallm}          & Assessing LLMs' accuracy and consistency in reflecting specific traits.                                                       & Personality-aligned LLM generation                                                                              & Correlate language features with assigned personality types effectively.                 & arXiv                                           &                                                                                       \\
\multirow{-4}{*}{Retention}       & \citep{Klapach2023comparative}       & Exploring emotional abilities of LLMs.                            & Emotional analysis of LLMs                                                                     & GPT4 scores the highest at 889.5/1000, while Bing AI scores the lowest at 427.8/1000 among the 5 LLMs for emotional capabilities.                                          & research archive of rising scholars             &                                                                                       \\\midrule
                                  & \citep{jia2023knowledge}             & Addressing challenges in emotional support conversations.                                                                                  & MODERN                                                                                                          & MODERN records the lowest PPL at 14.99, highest BLEU-1 score at 23.19, BLEU-2 at 10.13, BLEU-3 at 5.53, BLEU-4 at 3.39, ROUGE-L at 20.86, METEOR at 9.26, and the highest CIDEr score at 30.08, outperforming all SOTA baselines on the ESConv dataset.                                                           & arXiv                                           &                                                                                       \\
                                  & \citep{jeong2023chatbot}             & Enhancing realism and consistency of responses from LLMs                          & Multi-sensory contextual response generation                                                                    & Model with all components achieves improved performance, scoring 0.118 in METEOR and 0.209 in Sentence BERT.                                                              & arXiv                                           & https://github.com
                                  \/srafsasm/InfoRichBot                  \\
                                  & \citep{zhong2023memorybank}          & Improving long-term memory in LLMs                                                         & MemoryBank                                                                                                      & Exhibiting strong capability for long-term companionship.                                         & arXiv                                           &                                                                                       \\
     \multirow{-5}{*}{Recall}                              & \citep{qian2023harnessing}           & Exploring the performance of LLMs in generating empathetic responses                                                                      & Semantically similar in-context learning, two-stage interactive generation, and combination with knowledge base & With ChatGPT using 5-shot in-context learning, the Dist-1 and Dist-2 scores reach 2.96 and 18.29, and BLEU-4 score is 2.65.                                                                                                 & arXiv                                           &                                                                                       \\
         & \citep{wake2023bias}                 & Assessing ChatGPT's performance in recognizing emotions from text.                                                                         & Emotion recognition with ChatGPT                                                                                & Emotion datasets vary, indicating potential bias and instability in LLMs.                                         & arXiv                                           &                                                                                       \\\midrule
                                  & \citep{tu2023characterchat}          & Providing emotional support to individuals with diverse personalities & Persona-driven social support conversation                                                              & The results show high effectiveness, with emotional improvement averaging 4.37, problem solving) at 3.30, and active engagement at 4.89 on a 5-point scale.  & arXiv                                           & https://github.com
                                  \/morecry/CharacterChat                 \\
                                  & \citep{qi2023supervised}             & Identifying cognitive distortions and suicidal thoughts.                                  & Psychological risk classification with LLMs                                                   & The fine-tuned GPT-3.5 achieves an F1 score of 78.45\%, an improvement of 11.5\% over the base model.                                                & research square                                 & https://github.com
                                  \/hongzhiQ
                                  \/SupervisedVsLLM
                                  \-EfficacyEval \\
       \multirow{-5}{*}{Problem-Solving}                           & \citep{zhu2024reading}      &  Enhancing LLMs' empathetic understanding to infer users' underlying goals and psychological needs.                                      &  Prompt engineering.                                                                                   &      reaching up to 85\% accuracy in inferring users' goals and fundamental psychological needs                                                                                                                              & arXiv       &                \\
& \citep{sajja2023artificial}          & Creating an smart assistant for personalized learning in higher education.                                 & AI-enabled Intelligent Assistant (AIIA)                                                    & The framework shows enhanced personalized and adaptive learning.                                   & arXiv                                           &                                                                                       \\
& \citep{lai2023psy}                   & Addressing the increasing demand for mental health support.                             & Psy-LLM                                                                                                         & The framework generates coherent and relevant answers effectively.                                                  & arXiv                                           &                                                                                       \\\midrule
                                  & \citep{trott2023large}               & Studying the impact of language exposure on understanding mental states.                                              & Language exposure hypothesis testing                                                                            & GPT-3's predictions are sensitive to characters' knowledge states, and the accuracy reaches 74.5\% and 73.4\% in two evaluations.                                                                                  & Cognitive Science                               & https://osf.io/hu865/                                    \\
                                  & \citep{gandhi2023understanding}      & Investigate the ToM reasoning capabilities of LLMs                                                                     & Causal template-based evaluation framework                                                                      & GPT4 mirrors human ToM, while other LLMs struggle in comparison.                                             & arXiv                                           & https://osf.io/qxj2s                                     \\
                                  & \citep{sap2022neural}                & Exploring social intelligence and ToM in modern NLP systems.                                                      & Instruction-tuned, RLFH, neural ToM                                                                             & GPT-4 performs only 60\% accuracy on the ToM questions                                       & arXiv                                           &                                                                                       \\
        \multirow{-12}{*}{Thinking}  
        & \citep{lee2024large}            &  Investigating LLMs' abilities to generate empathetic responses                                                                                       & Human evaluation                                                                                            & GPT-4 achieves an average empathy score of 4.1 out of 5                                  & arXiv                                            &        \\
                                  & \citep{shapira2023clever}            & Evaluating the extent of N-ToM capabilities in LLMs                                                                                       & N-ToM evaluation                                                                                            & LLMs have ToM abilities but lack robust performance.                                  & arXiv                                           & https://github.com
                                  \/salavi/Clever\_Hans\_or
                                  \_N-ToM        \\
                                  & \citep{zhou2023far}                  & Evaluating LLMs' ability to link mental states with social actions.                                        & Thinking for Doing (T4D), Zero-shot prompting framework, Foresee and Reflect (FaR)                                                                                        & FaR boosts GPT-4's performance from 50\% to 71\% on T4D                                         & arXiv                                           &                                                                                       \\
                                  & \citep{sorin2023large}      &  Reviewing LLMs' empathy in healthcare.                                      &  A literature search on MEDLINE                                                                                   &      ChatGPT-3.5 outperforms humans in empathy-related tasks, achieving up to 85\%                                                                                                                              & MedRxiv  &                 \\
                                    & \citep{schaaff2023exploring}      &  Exploring ChatGPT's ability to understand and express empathy                                      &  A series of tests and questionnaires                                                                                   &      ChatGPT accurately identified emotions in 91.7\% of cases and produced parallel emotional responses in 70.7\% of conversations                                                                                                                              & arXiv  &                 \\
                                    & \citep{saito2023chatgpt}      &  Enhancing LLMs' empathetic dialogue speech synthesis                                       &  Use ChatGPT to extract context words and train an empathetic speech synthesis model.                                                                                   &      ChatGPT-EDSS achieves 3.52 for naturalness and 3.24 for speaking style similarity                                                                                                                              & INTERSPEECH 2023  & 
                                    \\
                                    
& \citet{holterman2023does} & understand whether ChatGPT-3 and ChatGPT-4 can exhibit ToM. & Using six well-known problems that focus on human reasoning biases and decision-making & ChatGPT-4 provides 30\% more correct answers than ChatGPT-3 even if it's based on incorrect assumptions or reasoning. &arXiv&\\
       & \citep{jin2022make}                  & Predicting human moral judgments for AI collaboration and safety.                                           & Moral Chain-of-Thought (MoralCoT) prompting strategy                                                            & MoralCoT outperforms existing LLMs by 6.2\% F1.                                                                   & NeurIPS                                    & https://huggingface.co
\/datasets/feradauto
\/MoralExceptQA  \\ 
& \citet{del2022empathy} & Enhancing GPT-3' ability to detect empathy in text. & Transfer learning and fine-tuning & Achieve an F1 score of 0.73 on the empathy detection task. &WASSA 2022&\\
\bottomrule 
\end{tabular}%
}
\caption{A list of representative papers among each level with motivation, approach, performance, venue, and code/datalink.}
\label{tab:paperlist2}
\end{table*}

\end{document}